%% file: main_ral.tex
\documentclass[letterpaper, 10pt, journal, twoside]{IEEEtran}
\usepackage{amsmath,amsfonts}
\usepackage{algorithmic}
\usepackage{algorithm}
\usepackage{array}
\usepackage[caption=false,font=normalsize,labelfont=sf,textfont=sf]{subfig}
\usepackage{textcomp}
\usepackage{stfloats}
\usepackage{url}
\usepackage{verbatim}
\usepackage{graphicx}
\usepackage{cite}
\usepackage{multirow}
\usepackage{xcolor}
\newcommand{\revised}[1]{\textcolor{black}{#1}}
\begin{document}

\title{Partially Observable Adversarial Patch Attacks on Vision-Language-Action Models in Robotics}

\markboth{IEEE Robotics and Automation Letters. Preprint Version. Accepted May 2026}%
{Wang \MakeLowercase{\textit{et al.}}: Partially Observable Adversarial Patch Attacks on VLA Models}

\author{Xiaofei Wang, Mingliang Han, Tianyu Hao, Yi Yang, Yun-Bo Zhao, and Keke Tang%
\thanks{Received 31 January 2026; Revised 31 March 2026; Accepted 27 May 2026. 
This paper was recommended for publication by Editor J\'ulia Borr\`as Sol upon evaluation of the Associate Editor and Reviewers' comments. 
This work was supported by the National Natural Science Foundation of China under Grant 62472117, the Deep Space Exploration Directed Cultivation Fund of USTC, the Guangdong Basic and Applied Basic Research Foundation under Grant 2025A1515010157, 
the Science and Technology Projects in Guangzhou under Grant 2025A03J0137, and the CCF-NetEase ThunderFire Innovation Research Funding under Grant CCF-Netease 202514. 
\emph{(Corresponding authors: Yun-Bo Zhao; Keke Tang.)}}%
\thanks{Xiaofei Wang is with the Department of Automation, University of Science and Technology of China, Hefei, China, and also with SmartMore Corporation 
(e-mail: wxf9545@mail.ustc.edu.cn).}%
\thanks{Mingliang Han is with the Department of Automation, University of Science and Technology of China, Hefei, China 
(e-mail: mlhan@mail.ustc.edu.cn).}%
\thanks{Tianyu Hao and Keke Tang are with the Cyberspace Institute of Advanced Technology, Guangzhou University, Guangzhou, China 
(e-mail: howty666@gmail.com; tangbohutbh@gmail.com).}%
\thanks{Yi Yang is with the 8th Medical Center of Chinese PLA General Hospital, Beijing, China 
(e-mail: sheepone0214@gmail.com).}%
\thanks{Yun-Bo Zhao is with the Department of Automation, University of Science and Technology of China, Hefei, China, and also with the Institute of Artificial Intelligence, Hefei Comprehensive National Science Center, Hefei, China 
(e-mail: ybzhao@ustc.edu.cn).}%
\thanks{Digital Object Identifier (DOI): see top of this page.}
}

\maketitle


\input{1-abstract}

\input{2-intro}

\input{3-related}

\input{4-formulation}

\input{5-method}

\input{6-experiment}

\input{7-conclusion}

\bibliographystyle{IEEEtran}
\bibliography{IEEEabrv,refs}

\end{document}

%% file: 1-abstract.tex
\begin{abstract}

Vision–language–action (VLA) models are gaining attention in robotics, yet their robustness to adversarial attacks remains largely unexplored.  
Existing work shows that adversarial patches can mislead VLA-based robots but assumes full access to the entire execution trajectory, an unrealistic requirement in practice.  
We address this limitation by formulating a partially observable threat model, where the adversary can exploit only a short prefix of the trajectory to generate a fixed patch applied to all subsequent frames.  
Under this setting, we propose a two-phase framework.  
First, we localize the patch using the model’s attention maps to identify visually critical regions that correspond to the full instruction.  
Then, we optimize the patch to disrupt the semantic grounding of target objects and increase the curvature of action trajectories, thereby compounding failures in both perception and control.
Extensive experiments in simulation and real-world robotic environments show that our method sustains  adversarial effects under partial observability, 
inducing long-horizon disruptions and significantly reducing task success rates.
\end{abstract}



%% file: 2-intro.tex
\section{Introduction}

\IEEEPARstart{V}{ision}-language-action (VLA) models have emerged as a powerful paradigm for robotic intelligence~\cite{kim2024openvla,black2024pi_0,bjorck2025gr00t}. By jointly integrating visual perception, natural language understanding, and action generation, they promise scalable generalist capabilities across diverse tasks and environments. As these models move toward deployment in domains such as service robotics~\cite{shi2020we} and manufacturing~\cite{guerin2015framework}, ensuring their safety and robustness is critical. Prior work in computer vision has shown that even imperceptible perturbations to input images can cause severe errors in deep models~\cite{szegedy2013intriguing,FGSM}, raising the question of how resilient VLA-based robots are to adversarial attacks and what risks such vulnerabilities may pose in practice.

Adversarial robustness of VLA models is still largely unexplored. To our knowledge, the most relevant prior study is by Wang \emph{et al.}~\cite{wang2024exploring}, who presented an empirical analysis showing that adding an adversarial patch to the visual input can mislead VLA-based robots. Their results highlight that even simple, physically realizable patches can severely disrupt task execution. However, this study does not provide a principled formulation of the threat model and assumes the adversary can optimize patches with access to the entire rollout, an idealized setting rarely attainable in practice.

These limitations motivate the need for a systematic treatment of adversarial attacks on VLAs. 
Notably, real-world adversaries are unlikely to observe entire trajectories; even when observation is possible, they can access at most a short prefix of the execution.
This raises the question of whether a single static patch, optimized only from early observations, can still induce long-term disruptive effects. 
Formulating this partially observable scenario is a key step toward understanding \revised{practical} vulnerabilities of VLA-based robots and guiding the development of more robust models.


\begin{figure}[!t]
    \centering
    \includegraphics[width=0.92\linewidth]{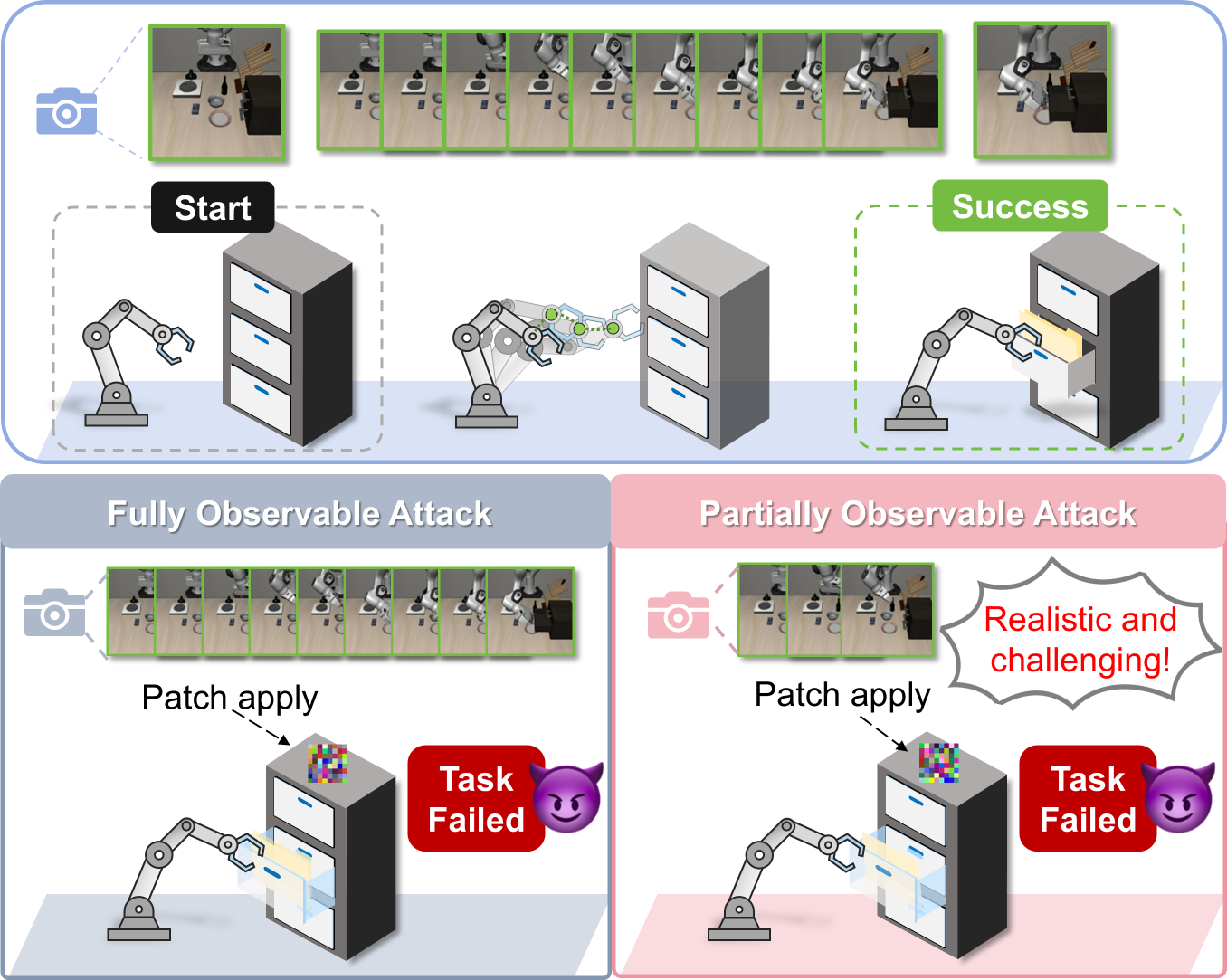}
\vspace{-1mm}
\caption{
Unlike attacks in fully observable settings that assume access to the entire trajectory, we study adversarial patch attacks on robotic VLA models under partial observability, where only the first few frames are available to design the patch. 
\revised{This setting is more constrained than full-rollout optimization and reflects limited-observation attack scenarios in robotic execution.}
}
\label{fig:teaser}
\end{figure}


In this paper, we formulate adversarial patch attacks on VLA models under a partially observable threat model, reflecting the practical constraint that an adversary can exploit only a limited prefix of the execution trajectory, see Fig.~\ref{fig:teaser}.  
\revised{This setting captures scenarios where an adversary can place a static physical patch in the robot-visible workspace but cannot observe or optimize over the full future rollout.}
Building on this formulation, we develop a two-phase attack.  
First, we localize the patch using the model’s attention maps, placing perturbations in visually critical regions that correspond to the instruction.  
We then optimize it to disrupt the visual grounding of instruction nouns and amplify trajectory curvature, disturbing smooth motion and reliable control.
Extensive experiments in simulation and on real robots show that our method sustains strong adversarial effects under partial observability, producing long-horizon disruptions that markedly reduce task success rates and outperform existing baselines.

Overall, our contribution is summarized as follows:
\begin{itemize}
\item We formulate fully and partially observable threat models for adversarial patch attacks on VLA models, with the latter reflecting   \revised{practical robotic constraints under limited observation}.
\item We propose a patch attack framework that exploits vulnerabilities in perception and control, disrupting semantic grounding and trajectory smoothness.

\item We validate the approach through extensive evaluation in simulation and on physical robots under practical conditions.

\end{itemize}

%% file: 3-related.tex
\section{Related Work}

\subsection{Security and Adversarial Attacks on Robots}

As robots are increasingly deployed in manufacturing, healthcare, and service domains~\cite{shi2020we,guerin2015framework}, their communication, perception, and control pipelines present attractive attack surfaces~\cite{yaacoub2022robotics}.  
Network attacks such as data injection or denial-of-service (DoS) can disrupt localization and planning~\cite{guerrero2018detection}, and even software frameworks such as the Robot Operating System (ROS) contain exploitable weaknesses~\cite{dieber2017security,mazzeo2020tros}.  
Beyond perception and control, motion planners are also vulnerable to adversarial attacks that mislead trajectory generation and degrade planning reliability~\cite{wu2024characterizing}.  
Adversarial manipulation has further been demonstrated in robotic grasping~\cite{alharthi2024physical,wang2019adversarial,wang2025advgrasp}, where subtle pixel-level or shape-based perturbations reduce grasp quality and stability, and in visual SLAM~\cite{299539}, where carefully designed patches compromise localization and mapping.  
These studies show that modern robotic systems remain exposed across networking, planning, and perception layers despite extensive conventional hardening.

\subsection{Attacks on VLA Models}

Security threats to VLA models are emerging beyond traditional perception attacks.  
Text-based jailbreaks adapt prompt-injection techniques from large language models to induce targeted robotic actions~\cite{jones2025adversarial}, and physical-world jailbreaks on embodied LLMs further reveal vulnerabilities in instruction following~\cite{zhang2024badrobot}.  
Decision-level robustness of LLM-based embodied agents has also been examined~\cite{liu2024exploring}, while backdoor attacks using objective-decoupled optimization expose another persistent threat vector~\cite{zhou2025badvla}.  
These studies show that VLA systems can be compromised through language-driven manipulation or hidden training-time triggers, even without direct visual perturbations.

Visual adversarial attacks on VLA models are less explored.  
Wang \emph{et al.}~\cite{wang2024exploring} demonstrate that carefully designed adversarial patches can mislead VLA-based robots, but their method assumes full rollout access and lacks a formal threat model, limiting practical applicability.  
In contrast, our work considers a more \revised{constrained partial-observation} setting where the adversary observes only a limited prefix of the trajectory and optimizes the patch within it, showing that disruptive visual attacks remain feasible under practical conditions.


%% file: 4-formulation.tex
\section{Problem Formulation}

\subsection{Problem Setup}
We consider a vision-language-action (VLA) model $f$ that maps a visual observation $x_t$ and a natural language instruction $l$ into an action $a_t$ in the robot’s control space:
\[
a_t = f(x_t, l), \quad t=1,\dots,T,
\]
where $x_t \in \mathbb{R}^{H \times W \times 3}$ denotes the image at time $t$, and $a_t \in \mathbb{R}^d$ is the executed action.  
The underlying system state $s_t \in \mathbb{R}^m$ evolves according to the transition dynamics
\[
s_{t+1} = \mathcal{T}(s_t, a_t),
\]
where $\mathcal{T}$ is the environment transition function.  
We denote by $\{a_t\}_{t=1}^T$ the clean action sequence and by $\{s_t\}_{t=1}^T$ the corresponding clean state trajectory.  

\revised{
The adversary follows a partial-access gray-box setting: it can query the VLA model and use model-side information required for patch optimization, but cannot modify model parameters, directly command robot actions, or alter the robot controller. 
The adversary can place a single static physical patch within the robot-visible scene. 
The patch belongs to a feasible set $\mathcal{P}$ that constrains its size and pixel intensity for physical realizability.
}

\subsection{Fully Observable Patch Attack Setting}

A common formulation assumes that the adversary has access to the entire observation sequence $\{x_t\}_{t=1}^T$ in advance.  
\revised{The patch is then optimized using all frames of the trajectory and, once learned, applied to every frame of the rollout.}
The resulting adversarial actions and states are
\[
\{\hat a_t\}_{t=1}^T = \{ f(x_t \oplus p, l) \}_{t=1}^T,
\qquad
\hat s_{t+1} = \mathcal{T}(\hat s_t, \hat a_t),
\]
where $x_t \oplus p$ denotes the observation frame $x_t$ with the patch $p$ superimposed.  
We denote by $\mathcal{L}_{adv}(\{s_t\},\{\hat s_t\})$ a task-level adversarial objective that measures the discrepancy between the clean states $\{s_t\}_{t=1}^T$ and the adversarial states $\{\hat s_t\}_{t=1}^T$.  
The optimization problem is therefore
\begin{equation}
p^{\mathrm{full}} = \arg\max_{p \in \mathcal{P}} \;
\mathcal{L}_{adv}\big(\{s_t\}_{t=1}^T, \{\hat s_t\}_{t=1}^T\big).
\end{equation}

This setting represents the strongest adversary with full observability.  
However, it presumes that the attacker can optimize $p$ with knowledge of the entire trajectory, which is unrealistic in real-world robotic deployments.  
\revised{
It thus serves mainly as a theoretical upper bound that motivates more practical alternatives.
}

\begin{figure*}[!t]
\vspace{2mm}
    \centering
    \includegraphics[width=0.99\linewidth]{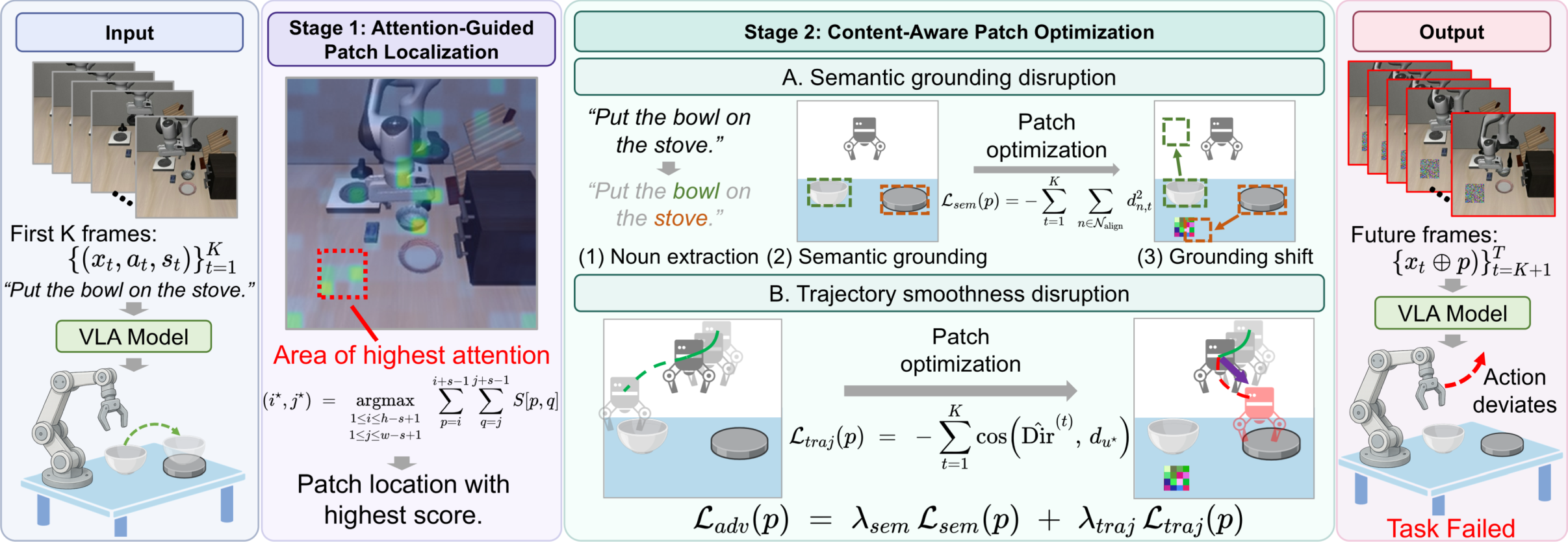}
    \vspace{-1mm}
    \caption{
Overview of the proposed attack framework under partial observability. 
In Stage 1, we perform attention-based patch localization to identify a compact region for adversarial patch placement.
In Stage 2, we perform content-aware patch optimization to alter object grounding and bias actions toward curvature-amplifying directions, leading to task failures.
    }
    \label{fig:framework}
\end{figure*}

\subsection{Partially Observable Patch Attack Setting}
To better capture practical deployments, we introduce a partially observable setting.  
Here the adversary can only observe a prefix of the trajectory, i.e., the first $K \ll T$ frames with their state-action tuples $\{(x_t,a_t,s_t)\}_{t=1}^K$.  
Based on this limited information, the adversary must design a \emph{single static patch} $p \in \mathcal{P}$, which is then fixed and applied to all subsequent frames $t=K+1,\dots,T$.  

\paragraph{Patch learning from the prefix}
The patch is optimized by maximizing an adversarial objective defined on the observable prefix:
\begin{equation}
\label{eq:po-train}
p^{\ast} = \arg\max_{p \in \mathcal{P}} \;
\mathcal{L}_{adv}\!\left(
\{s_t\}_{t=1}^K,\,
\{\hat s_t\}_{t=1}^K
\right),
\end{equation}
where $\{s_t\}_{t=1}^K$ are the clean states and $\{\hat s_t\}_{t=1}^K$ are the adversarial states produced by patched prefix observations.  
Although optimization is performed only on the prefix, the learned patch is intended to generalize and degrade the unseen future trajectory.  

\paragraph{Evaluation on the unseen future}
Once $p^{\ast}$ is fixed, it is applied to all subsequent frames $t=K+1,\dots,T$.  
This produces adversarial actions
\[
\{\hat a_t\}_{t=K+1}^T = \{ f(x_t \oplus p^{\ast}, l) \}_{t=K+1}^T,
\]
and the corresponding state rollout
\[
\hat s_{K} = s_{K}, 
\qquad 
\hat s_{t+1} = \mathcal{T}(\hat s_t, \hat a_t), \quad t=K+1,\dots,T.
\]
The attack effectiveness is then \revised{evaluated on the unseen suffix as}
\begin{equation}
\label{eq:po-eval}
\revised{
\mathcal{E}_{\mathrm{suffix}}(p^{\ast})
=
}
\mathcal{L}_{adv}\!\left(
\{s_t\}_{t=K+1}^T,
\{\hat s_t\}_{t=K+1}^T
\right),
\end{equation}
where higher values indicate stronger adversarial degradation on the unseen suffix trajectory.

%% file: 5-method.tex
\section{Methodology}

To realize adversarial patch attacks under the partially observable setting, 
we propose a two-phase framework: (i) \emph{attention-based patch localization}, 
which identifies a compact and influential region, and (ii) \emph{content-aware patch optimization}, 
which disrupts semantic grounding and trajectory smoothness. 
Please refer to Fig.~\ref{fig:framework} for demonstration.

\subsection{Attention-Based Patch Localization}
\label{sec:localization}

To determine an effective patch location, we leverage cross-modal attention from the VLA model.
Following the standard ViT backbone \cite{Dosovitskiy-ViT}, each frame is encoded into $N_v = h \times w$ patch tokens arranged on a 2D grid.
We compute attention using the last available prefix frame $x_K$, as it provides the most recent visual context before execution.
The instruction $l$ is tokenized into $Z$ text tokens $\{w_z\}_{z=1}^Z$, and the model generates a head-summed attention matrix $A^{(K)} \in \mathbb{R}^{Z \times N_v}$.

We define the reshape operator that maps a length-$N_v$ vector back to the $h\times w$ token grid:
\begin{equation}
\mathrm{Reshape}_{h\times w} : \mathbb{R}^{N_v} \;\to\; \mathbb{R}^{h\times w}.
\label{eq:reshape-def}
\end{equation}

We then aggregate across text tokens by taking the elementwise maximum at each grid location. 
Specifically, the saliency map $S \in \mathbb{R}^{h \times w}$ is defined as
\begin{equation}
S[i,j] \;=\; \max_{1 \le z \le Z}
\Big(\mathrm{Reshape}_{h\times w}(A^{(K)}_{z,:})\Big)[i,j],
\label{eq:saliency}
\end{equation}
for all $i=1,\dots,h$ and $j=1,\dots,w$, where $[\cdot,\cdot]$ denotes the matrix entry.

To constrain attacks to a compact contiguous region, we slide an $s\times s$ window over $S$ and select the top-left index that maximizes the aggregated score:
\begin{equation}
(i^\star,j^\star)
\;=\;
\operatorname*{argmax}_{\substack{1 \le i \le h-s+1 \\[2pt] 1 \le j \le w-s+1}}
\;\sum_{p=i}^{i+s-1}\sum_{q=j}^{j+s-1} S[p,q].
\label{eq:window-argmax}
\end{equation}
The window anchored at $(i^\star,j^\star)$ defines the feasible patch region.
We denote by $M_{\mathrm{patch}} \in \{0,1\}^{H\times W}$ the binary mask at image resolution, 
obtained by upsampling the selected $h \times w$ token window to the original image size. 
This mask restricts subsequent optimization updates to the region.

\subsection{Content-Aware Patch Optimization}
\label{sec:optimization}

Given the fixed patch location, we optimize its content to maximize adversarial impact. 
The overall loss $\mathcal{L}_{adv}$ comprises a semantic term that shifts language--vision grounding of target objects and a trajectory term that promotes irregular robot motion, with all updates to $p$ restricted by $M_{\mathrm{patch}}$.

\paragraph{Semantic grounding disruption}
Robotic policies rely on correctly grounding task-relevant nouns in the instruction to corresponding visual regions. 
If this grounding is shifted, errors can propagate into downstream actions and states. 
\revised{We use attention displacement as an operational proxy for changes in language--vision grounding, rather than a direct causal explanation of the model decision.}

Let $\mathcal{N}_{\mathrm{align}}$ denote the set of task-relevant nouns extracted from the instruction. 
Each noun $n \in \mathcal{N}_{\mathrm{align}}$ is associated with a token span $S_n \subseteq \{1,\dots,Z\}$ corresponding to the subword tokens forming this noun. 
For each prefix frame index $t \in \{1,\dots,K\}$, 
let $\hat A^{(t)}$ denote the attention matrix corresponding to $A^{(t)}$ when the adversarial patch is applied. 
We then construct the clean and adversarial grounding maps by reshaping and aggregating these scores:
\begin{equation}
\begin{aligned}
M^{(t)}_n[i,j] \;&=\; \sum_{z \in S_n}
   \Big(\mathrm{Reshape}_{h\times w}(A^{(t)}_{z,:})\Big)[i,j], \\[4pt]
\hat M^{(t)}_n[i,j] \;&=\; \sum_{z \in S_n}
   \Big(\mathrm{Reshape}_{h\times w}(\hat A^{(t)}_{z,:})\Big)[i,j].
\end{aligned}
\label{eq:grounding-map}
\end{equation}

The clean focus $(x_{n}^{t}, y_{n}^{t})$ and the adversarial focus $(\hat{x}_{n}^{t}, \hat{y}_{n}^{t})$ 
are obtained via soft-argmax over $M_n^{(t)}$ and $\hat M_n^{(t)}$, respectively. 
The displacement is then measured as
\begin{equation}
d_{n,t}^2 \;=\; (\hat{x}_{n}^{t} - x_{n}^{t})^2 + (\hat{y}_{n}^{t} - y_{n}^{t})^2,
\label{eq:displacement}
\end{equation}
and the semantic loss is defined by
\begin{equation}
\mathcal{L}_{sem}(p) \;=\; - \sum_{t=1}^{K}\ \sum_{n \in \mathcal{N}_{\mathrm{align}}} d_{n,t}^2,
\label{eq:Lsem}
\end{equation}
which encourages large shifts in the attention-derived grounding focus of target objects across the observed prefix.

\paragraph{Trajectory smoothness disruption}
Nominal robot motion commonly follows smoothness preferences, such as low-curvature and low-jerk motion in manipulation and trajectory planning\revised{~\cite{Flash1985Coordination,Gasparetto2008TimeJerkOptimal}. 
Motivated by this prior, we use curvature increase as a practical surrogate for inducing irregular motion, rather than as a direct measure of task failure.}
Our strategy is to identify the direction that amplifies curvature relative to the prefix trajectory and steer the VLA outputs under perturbation toward it.

Let $\{P_t\}_{t=1}^{K}$ be the end-effector positions from the prefix (the position component of $\{s_t\}_{1:K}$).  
We sample $U$ candidate future directions $\{d_u\}_{u=1}^U$ uniformly on the unit sphere and normalize their length to match the typical step scale. 
For a horizon $T_f$, each direction yields a stitched sequence
\begin{equation}
\mathcal{P}^{(u)}_{\mathrm{full}} = \{P_1,\dots,P_{K},\, P_{K}+d_u,\dots,P_{K}+T_f d_u\}.
\label{eq:stitched}
\end{equation}
Since observations are discrete, we fit a cubic spline $\mathbf{r}^{(u)}(t)=(x^{(u)}(t),y^{(u)}(t),z^{(u)}(t))$ through $\mathcal{P}^{(u)}_{\mathrm{full}}$ to obtain a smooth continuous representation suitable for differential geometry analysis. 
The curvature of the fitted trajectory is
\begin{equation}
\kappa^{(u)}(t) \;=\; \frac{\|\mathbf{r}^{(u)\prime}(t)\times \mathbf{r}^{(u)\prime\prime}(t)\|}{\|\mathbf{r}^{(u)\prime}(t)\|^3}.
\label{eq:curvature}
\end{equation}
Let $\bar\kappa_{\mathrm{hist}}$ be the maximum curvature over the prefix spline, and $\bar\kappa^{(u)}_{\mathrm{full}}$ that of the stitched spline. 
We identify the adversarial target direction as
\begin{equation}
u^\star \;=\; \arg\max_{1 \le u \le U}\ \Big(\bar\kappa^{(u)}_{\mathrm{full}} - \bar\kappa_{\mathrm{hist}}\Big),
\label{eq:worst-dir}
\end{equation}
which corresponds to the sampled direction with the largest curvature increase.

Let $\mathrm{Dir}^{(t)}$ and $\hat{\mathrm{Dir}}^{(t)}$ denote the action directions predicted by the VLA at prefix frame $t$ under the clean and adversarial settings, respectively. 
The trajectory loss encourages the adversarial directions to align with the selected high-curvature direction $d_{u^\star}$:
\begin{equation}
\mathcal{L}_{traj}(p) \;=\; - \sum_{t=1}^K \cos\!\Big(\hat{\mathrm{Dir}}^{(t)},\, d_{u^\star}\Big).
\label{eq:Ltraj}
\end{equation}

\paragraph{Unified adversarial objective}
The final adversarial objective integrates both components:
\begin{equation}
\mathcal{L}_{adv}(p) \;=\; \lambda_{sem}\,\mathcal{L}_{sem}(p) \;+\; \lambda_{traj}\,\mathcal{L}_{traj}(p).
\label{eq:Ladv}
\end{equation}
The patch $p$ is optimized on the prefix data $\{(x_t,a_t,s_t)\}_{t=1}^K$, with gradient updates restricted by $M_{\mathrm{patch}}$.

%% file: 6-experiment.tex
\section{Experiments}

\subsection{Experimental Setup}


\paragraph{Implementation}
We implement the attack in PyTorch and optimize it with SGD using an exponential learning rate from 20.0, decaying by 0.95 per iteration over 50 iterations. 
The prefix length $K$ defaults to 40 unless noted otherwise. 
During localization, we select the contiguous token window with the highest saliency on the last prefix frame, with default size $s{=}3$ (i.e., $42\times 42$ pixels). 
The loss weights in Eq.~\eqref{eq:Ladv} are $\lambda_{sem}{=}1.0$ and $\lambda_{traj}{=}12.0$, and \revised{the number of candidate directions is $U{=}100$ by default}. 
All experiments run on a workstation with eight H100 GPUs.


\paragraph{Benchmark and Victim VLAs}
All evaluations are performed on the LIBERO benchmark~\cite{liu2023libero}, 
which contains four task categories: Spatial, Object, Goal, and Long. 
Each category includes 10 tasks, and every task is executed for 100 trials, yielding a total of 1{,}000 rollouts. 
\revised{The maximum rollout horizons are 220, 280, 300, and 520 steps for Spatial, Object, Goal, and Long, respectively.}
As victim models, we attack four OpenVLA~\cite{kim2024openvla} variants, each trained on one suite, and evaluate Hume~\cite{song2025hume} with official checkpoints fine-tuned on the same suites to test a different architecture.

\paragraph{Baselines}
We compare our method with three patch attack methods introduced in \cite{wang2024exploring}: \textsc{UADA}, \textsc{UPA}, and \textsc{TMA}. 
All baselines use the same prefix length $K$ and number of optimization iterations as ours for fair comparison.

\begin{table*}[!t]
\setlength\tabcolsep{5pt}
\centering

\caption{
Attack success rate (ASR, \%) of UADA, UPA, TMA, and Ours on the LIBERO benchmark
across the four suites (Spatial, Object, Goal, Long) under prefix lengths $K\in\{10,20,30,40\}$.
}
\vspace{-2mm}
\label{tab:asr}
\begin{tabular}{c|cccc|cccc|cccc|cccc}

\hline
                              & \multicolumn{4}{c|}{Spatial}                                  & \multicolumn{4}{c|}{Object}                                   & \multicolumn{4}{c|}{Goal}                                     & \multicolumn{4}{c}{Long}                                      \\ \cline{2-17} 
                              & K=10          & K=20          & K=30          & K=40          & K=10          & K=20          & K=30          & K=40          & K=10          & K=20          & K=30          & K=40          & K=10          & K=20          & K=30          & K=40          \\ \hline
UADA                          & 21.1          & 53.6          & 57.4          & 46.1          & 32.6          & 35.9          & 63.8          & 61.9          & 28.6          & 32.4          & 59.5          & 59.4          & 12.0          & 47.5          & 54.7          & 58.5          \\
UPA                           & \textbf{35.2} & 49.3          & 53.8          & 48.0          & 34.2          & 41.8          & 59.7          & 57.9          & 29.6          & 37.3          & 53.0          & 55.3          & 37.3          & 42.7          & 57.2          & 59.1          \\
TMA                           & 33.2          & 51.4          & 59.1          & 52.6          & 31.4          & 37.7          & 62.1          & 61.4          & 35.3          & 36.9          & 54.5          & 57.7          & 32.3          & 45.1          & 44.4          & 47.4          \\
Ours                          & 33.4          & \textbf{70.4} & \textbf{73.8} & \textbf{72.4} & \textbf{83.2} & \textbf{86.4} & \textbf{90.7} & \textbf{89.7} & \textbf{47.0} & \textbf{60.2} & \textbf{72.8} & \textbf{71.7} & \textbf{71.0} & \textbf{73.8} & \textbf{86.6} & \textbf{89.1}
\\ \hline
\end{tabular}
\end{table*}

\begin{table*}[!t]
\setlength\tabcolsep{5pt}
\centering

\caption{
Normalized attack success rate (nASR, \%) of UADA, UPA, TMA, and Ours on the LIBERO benchmark
across the four suites (Spatial, Object, Goal, Long) under prefix lengths $K\in\{10,20,30,40\}$.
}
\vspace{-2mm}
\label{tab:ourmetrics}
\begin{tabular}{c|cccc|cccc|cccc|cccc}

\hline
                              & \multicolumn{4}{c|}{Spatial}                                  & \multicolumn{4}{c|}{Object}                                   & \multicolumn{4}{c|}{Goal}                                     & \multicolumn{4}{c}{Long}                                      \\ \cline{2-17} 
                              & K=10          & K=20          & K=30          & K=40          & K=10          & K=20          & K=30          & K=40          & K=10          & K=20          & K=30          & K=40          & K=10          & K=20          & K=30          & K=40          \\ \hline
UADA                          & 59.9          & 73.2          & 77.7          & 72.2          & 63.8          & 67.0          & 82.3          & 80.0          & 56.0          & 57.1          & 74.4          & 75.8          & 73.3          & 82.2          & 84.9          & 85.0          \\
UPA                           & 67.4          & 69.9          & 74.5          & 72.2          & 65.7          & 69.7          & 79.1          & 78.2          & 55.0          & 60.4          & 71.0          & 69.4          & 77.4          & 73.1          & 81.7          & 82.3          \\
TMA                           & 65.4          & 65.2          & 73.4          & 74.1          & 63.0          & 68.9          & 80.6          & 79.7          & 59.1          & 59.0          & 69.3          & 73.7          & 74.3          & 74.6          & 79.9          & 81.8          \\
Ours                          & \textbf{67.6} & \textbf{86.2} & \textbf{87.5} & \textbf{86.9} & \textbf{92.8} & \textbf{93.7} & \textbf{96.0} & \textbf{95.2} & \textbf{67.2} & \textbf{75.7} & \textbf{79.6} & \textbf{83.9} & \textbf{92.8} & \textbf{92.7} & \textbf{93.9} & \textbf{94.8}\\
\hline
\end{tabular}
\end{table*}

\paragraph{Evaluation metrics}
We assess attack performance using two metrics:

\begin{itemize}[]
  \item \textbf{ASR} (Attack Success Rate): the fraction of episodes that fail under attack according to the LIBERO task-specific success criterion, i.e., one minus the success rate under attack.
  \item \textbf{nASR} (Normalized Attack Success Rate): a unified score equal to $1$ when the attack causes task failure and a proportional value in $(0,1)$ when it only induces delay. 
  For rollout $r$,
\[
\mathrm{nASR}(r)=
\begin{cases}
1, & \text{task fails},\\[1pt]
\dfrac{\max\!\bigl(0,\,T_{\text{attack}}(r)-T_{\text{benign}}(r)\bigr)}
      {T_{\max}(r)+1-T_{\text{benign}}(r)}, & \text{else}.
\end{cases}
\]
Here $T_{\text{benign}}(r)$ is the completion steps of the benign rollout, $T_{\text{attack}}(r)$ the completion steps under attack when the task completes, and $T_{\max}(r)$ the maximum steps allowed for task $r$ in LIBERO.\end{itemize}

\subsection{Main Results}

\paragraph{Comparison with state-of-the-art methods}
Tab.~\ref{tab:asr} reports ASR on the four LIBERO suites under different prefix lengths. Our method achieves the highest attack success in almost all settings, consistently outperforming all baselines. The advantage is most evident at a moderate prefix (e.g., $K{=}30$), which offers enough context to locate semantically critical regions and optimize an effective patch within the fixed budget. UADA, UPA, and TMA struggle under short prefixes because updates fitted on limited observations cannot capture the evolving scene, with the largest gap on Goal and Long. 

Tab.~\ref{tab:ourmetrics} reports nASR, which also credits rollout slowdowns. Our method again performs best in nearly every configuration, delaying even successful rollouts by inducing less efficient trajectories.

\begin{figure*}[!t]
    \centering
    \includegraphics[width=0.9999\linewidth]{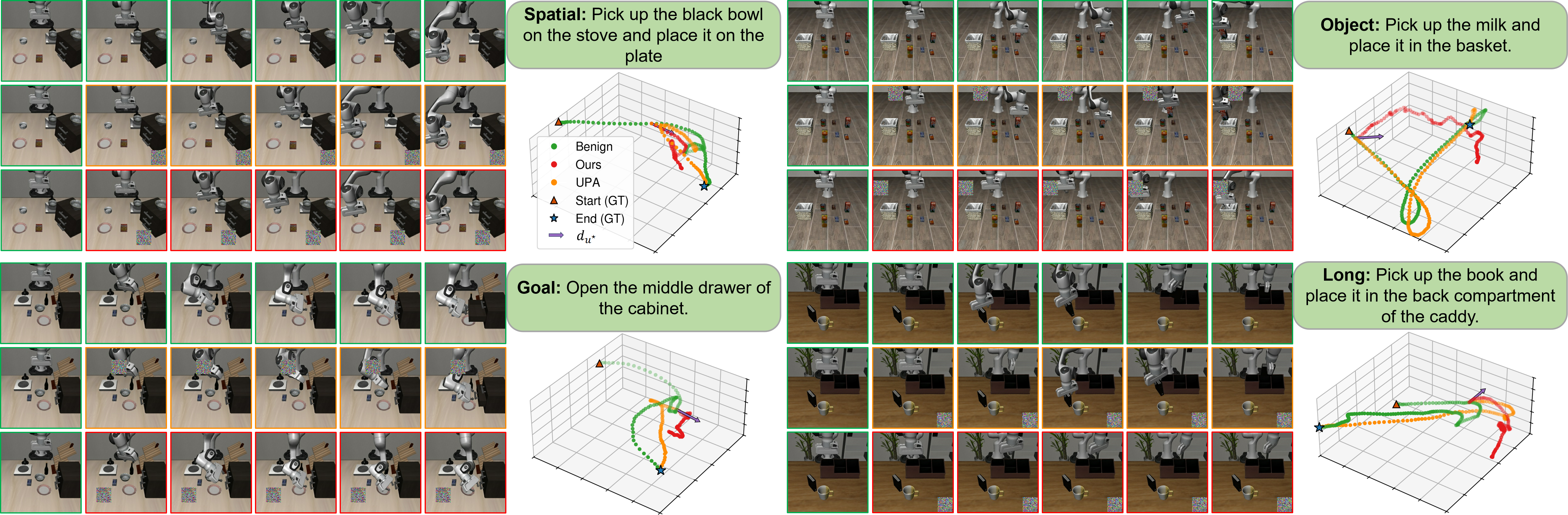}
\vspace{-1mm}
\caption{
Snapshots of OpenVLA executions on four LIBERO tasks under different attack settings.  
The top row shows benign execution, the middle row shows execution attacked by UPA, and the bottom row shows execution with our adversarial patch; both attacks are applied with prefix length $K{=}30$.  
The rightmost panels depict the 3D gripper trajectories for each setting, with start and goal markers and a purple arrow indicating the high-curvature direction selected by our attack.
}
    \label{fig:curvature}
\end{figure*}

\paragraph{Visualization of task executions under adversarial attack}
\label{sec:curvature}
Fig.~\ref{fig:curvature} presents representative executions of OpenVLA on four LIBERO tasks under three settings: benign, attacked by UPA, and attacked by our method, with both attacks optimized at $K{=}30$.  
When our patch is applied, the policy rapidly deviates from the nominal path and, after brief recovery attempts, is redirected along high-curvature branches.  
\revised{This behavior is consistent with the roles of the two loss terms: $\mathcal{L}_{sem}$ alters noun-related grounding behavior, while $\mathcal{L}_{traj}$ biases motion toward curvature-amplifying directions.}
Under partial observability these effects cause oscillations, overshoots, and detours that repeatedly trap corrective actions in high-curvature regions and make task completion difficult.

In contrast, UPA induces only transient deviations from which the robot typically recovers. Suite-specific patterns are also evident: sharp bends near waypoints in Spatial and Object, failure to stabilize for grasp or placement in Goal, and accumulated deviations with frequent timeouts in Long, consistent with the ASR and nASR trends.

\subsection{Ablation Studies and Additional Analyses}
\label{sec:ablation}

\begin{figure}[!t]
    \centering
    \includegraphics[width=0.7\linewidth]{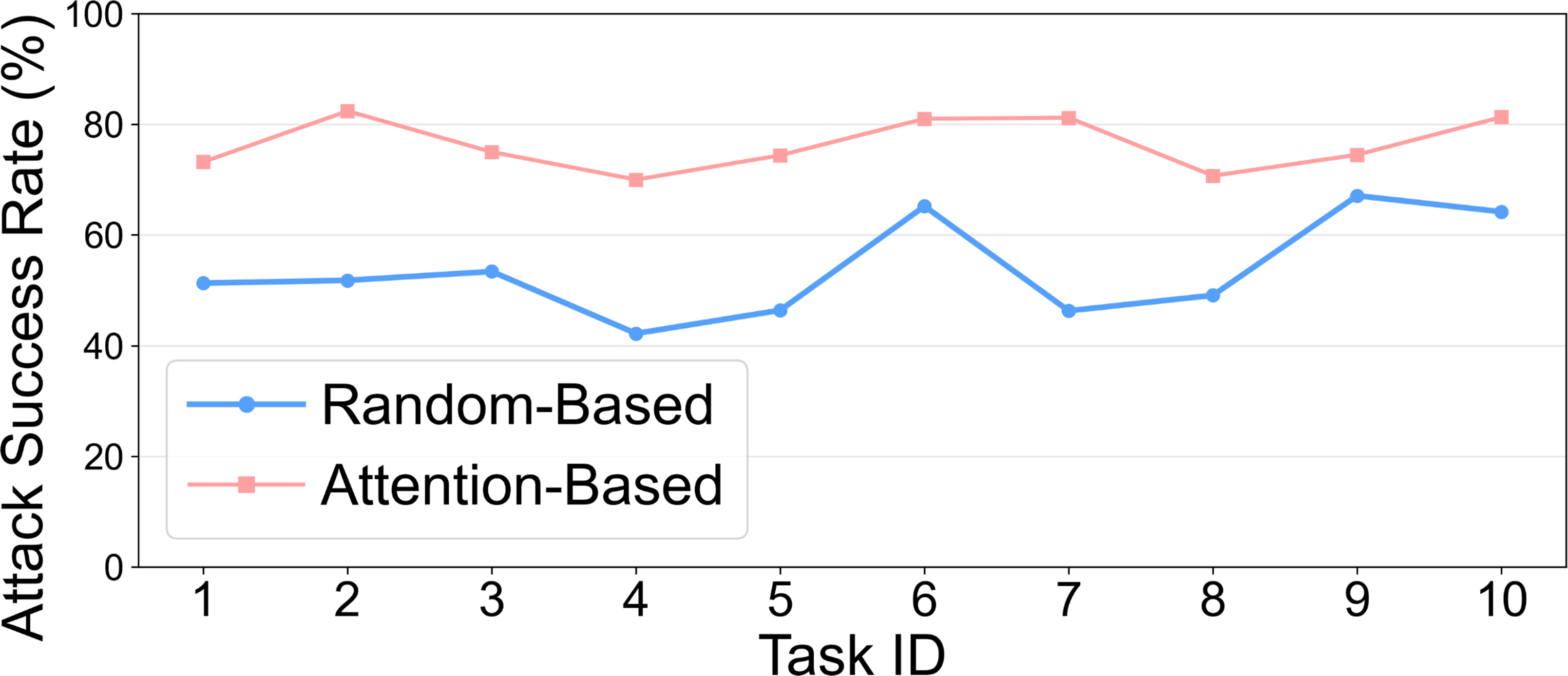}
\vspace{-1mm}
\caption{
Attack success rate (ASR, \%) of our attack framework with random and attention-based patch location on ten LIBERO–Goal tasks.
}
  \label{fig:location}
\end{figure}

\paragraph{\revised{Importance of attention-based patch localization}}
\revised{
To examine the role of patch localization, we compare attention-based placement with random placement on the LIBERO-Goal suite while keeping all other attack parameters fixed. 
Fig.~\ref{fig:location} shows that attention-based localization consistently improves ASR, confirming the benefit of placing patches in semantically important regions.
A broader comparison against stronger localization heuristics is left for future work.}

\revised{
We further compare two attention aggregation strategies: the last prefix frame $x_K$ versus averaging over all prefix frames. 
As shown in Tab.~\ref{tab:localization_strategy}, $x_K$ achieves better average ASR/nASR, especially at $K{=}30$, supporting our use of the most recent observation.
}

\begin{table}[!t]
\setlength\tabcolsep{5pt}
\centering
\caption{\revised{
Average ASR/nASR (\%) for patch localization strategies, averaged over the four LIBERO suites.
\emph{Prefix Mean} selects the patch location using attention averaged over all prefix frames, while \emph{Last Frame} selects it using only $x_K$.
}}
\label{tab:localization_strategy}
\vspace{-2mm}
\begin{tabular}{c|l|cc}
\hline
\revised{$K$} & \revised{Strategy} & \revised{ASR} & \revised{nASR} \\ \hline
\multirow{2}{*}{\revised{10}} 
& \revised{Prefix Mean} & \revised{58.0} & \revised{75.8} \\
& \revised{Last Frame}  & \revised{\textbf{58.7}} & \revised{\textbf{80.1}} \\ \hline
\multirow{2}{*}{\revised{30}} 
& \revised{Prefix Mean} & \revised{77.7} & \revised{86.6} \\
& \revised{Last Frame}  & \revised{\textbf{81.0}} & \revised{\textbf{89.3}} \\ \hline
\end{tabular}
\vspace{-2mm}
\end{table}

\begin{figure}[!t]
    \centering
    \vspace{1mm}
    \includegraphics[width=0.99\linewidth]{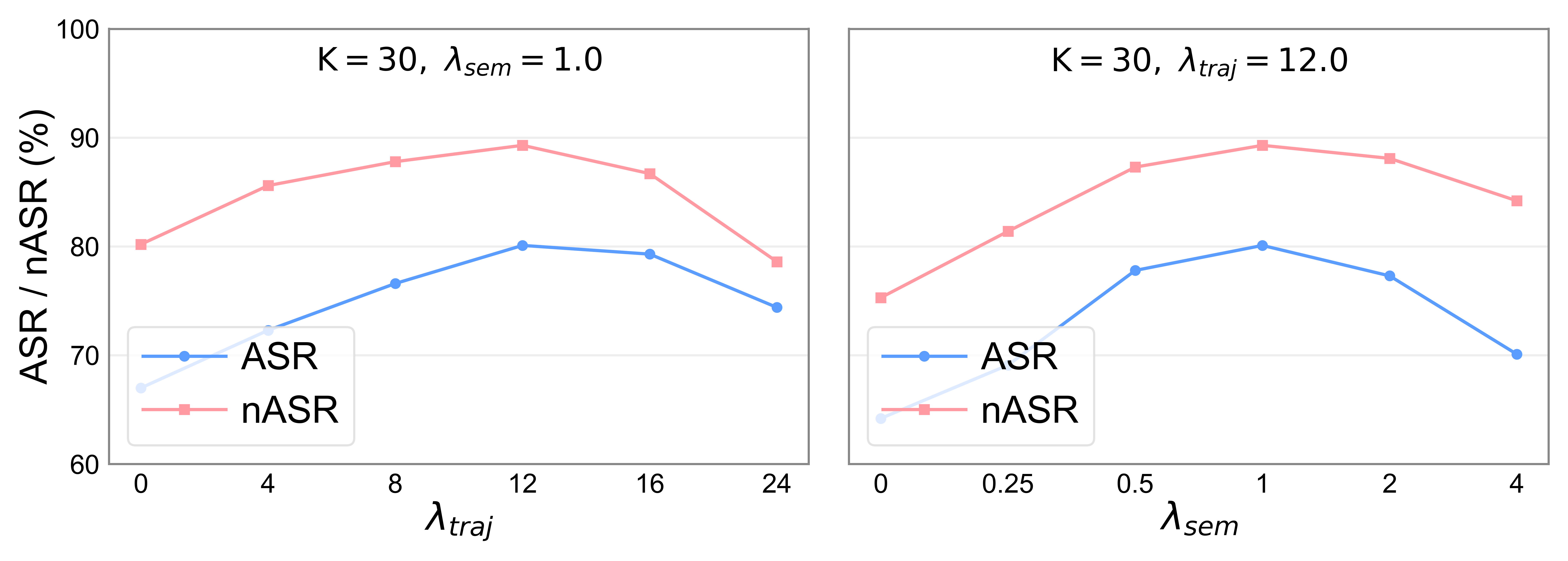}
    \vspace{-2mm}
\caption{\revised{
Effect of loss weights on ASR and nASR.
Left: varying $\lambda_{traj}$ with $\lambda_{sem}{=}1.0$; right: varying $\lambda_{sem}$ with $\lambda_{traj}{=}12.0$.
}}
    \label{fig:lambda_sensitivity}
\end{figure}

\paragraph{\revised{Effect of semantic and trajectory losses}}
\revised{
To assess the contribution of the two loss terms, we ablate each component individually.
Tab.~\ref{tab:ablation} reports ASR and nASR across LIBERO suites at $K{=}10$ and $30$: removing either $\mathcal{L}_{sem}$ or $\mathcal{L}_{traj}$ degrades performance, while the full model achieves the best results across most settings, indicating that both terms contribute.
We further examine the loss weights in Eq.~\eqref{eq:Ladv}; as shown in Fig.~\ref{fig:lambda_sensitivity}, moderate weights for both terms yield stronger attacks, and the default $(\lambda_{sem},\lambda_{traj})=(1,12)$ achieves the best nASR with near-best ASR.
}

\begin{table}[!t]
\setlength\tabcolsep{1.8pt}
\centering
\vspace{-2mm}
\caption{
Attack success rate (ASR, \%) and normalized attack success rate (nASR, \%) of our attack framework on representative tasks from the four LIBERO suites with prefix lengths $K\in\{10,30\}$.
\revised{Results are reported for the full model and variants without the semantic loss $\mathcal{L}_{sem}$ or trajectory loss $\mathcal{L}_{traj}$.}
}
\label{tab:ablation}
\vspace{-2mm}
\begin{tabular}{cc|cc|cc|cc|cc}
\hline
                                           &                               & \multicolumn{2}{c|}{Spatial}  & \multicolumn{2}{c|}{Object}   & \multicolumn{2}{c|}{Goal}     & \multicolumn{2}{c}{Long}      \\ \cline{3-10} 
                                           &                               & K=10          & K=30          & K=10          & K=30          & K=10          & K=30          & K=10          & K=30          \\ \hline
\multicolumn{1}{c|}{\multirow{3}{*}{ASR}}  &  w/o $\mathcal{L}_{sem}$  & 29.9          & 60.8          & 73.3          & 78.9          & \textbf{49.2} & 58.5          & 37.2          & 58.5          \\
\multicolumn{1}{c|}{}                      &  w/o $\mathcal{L}_{traj}$ & \textbf{34.9} & 63.6          & 64.2          & 72.1          & 33.9          & 59.4          & 54.5          & 73.0          \\
\multicolumn{1}{c|}{}                      & Full                          & 33.4          & \textbf{73.8} & \textbf{83.2} & \textbf{90.7} & 47.0 & \textbf{72.8} & \textbf{71.0} & \textbf{86.6} \\ \hline
\multicolumn{1}{c|}{\multirow{3}{*}{nASR}} &  w/o $\mathcal{L}_{sem}$  & \textbf{67.8} & 81.3          & 87.5          & 89.4          & \textbf{68.5} & 74.8          & 84.7          & 83.6          \\
\multicolumn{1}{c|}{}                      &  w/o $\mathcal{L}_{traj}$ & 66.4          & 82.6          & 82.7          & 86.1          &   58.9   & 73.5          & 85.1          & 90.7          \\
\multicolumn{1}{c|}{}                      & Full                          & 67.6          & \textbf{87.5} & \textbf{92.8} & \textbf{96.0} & 67.2          & \textbf{79.6} & \textbf{92.8} & \textbf{93.9} \\ \hline
\end{tabular}
\vspace{-2mm}
\end{table}

\paragraph{\revised{Computational cost}} 
\revised{Tab.~\ref{tab:cost} reports patch generation time at $K{=}30$ on a single NVIDIA H100 GPU. 
Compared with UADA-style full-rollout optimization, our attack reduces the cost from 351--783\,s to 47--53\,s across suites, a 6.9$\times$--16.7$\times$ speedup.}

\begin{table}[!t]
\setlength\tabcolsep{5pt}
\centering
\caption{\revised{
Patch generation time (s) at $K{=}30$ on a single NVIDIA H100 GPU.
\emph{Full Rollout} denotes UADA-style optimization over the complete trajectory. Lower is better.
}}\label{tab:cost}
\vspace{-2mm}
\begin{tabular}{l|cccc}
\hline
\revised{Method} & \revised{Spatial} & \revised{Object} & \revised{Goal} & \revised{Long} \\ \hline
\revised{Full Rollout} & \revised{351} & \revised{413} & \revised{492} & \revised{783} \\
\revised{Ours} & \revised{51} & \revised{48} & \revised{53} & \revised{47} \\ \hline
\end{tabular}
\vspace{-2mm}
\end{table}



\begin{figure}[!t]
    \centering
\vspace{1mm}
    \includegraphics[width=0.99\linewidth]{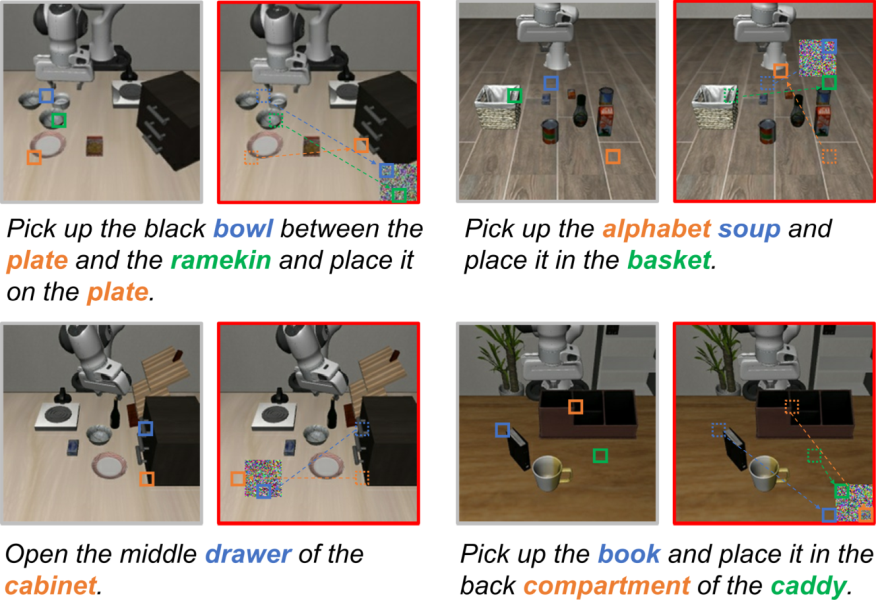}
\vspace{-1mm}
\caption{
Visual grounding maps of instruction nouns before and after our adversarial attack on four LIBERO tasks.
Each pair shows the benign frame (left) and the attacked frame at prefix length $K{=}30$ (right, red border), with colored boxes marking peak grounding responses for the corresponding nouns.
}
    \label{fig:attention}
\end{figure}

\paragraph{\revised{Analysis of semantic grounding disruption}}
\revised{
We evaluate $\mathcal{L}_{sem}$ by visualizing grounding maps for instruction nouns before and after applying our patch at $K{=}30$ in Fig.~\ref{fig:attention}.
In benign executions, the maps concentrate on the intended referents (e.g., the bowl and the book); after the patch is applied, the peaks shift away toward background regions or areas adjacent to the patch.
This provides attention-based evidence that $\mathcal{L}_{sem}$ alters noun-related grounding, consistent with degraded execution.
}

\begin{figure}[!t]
    \centering
    \includegraphics[width=0.83\linewidth]{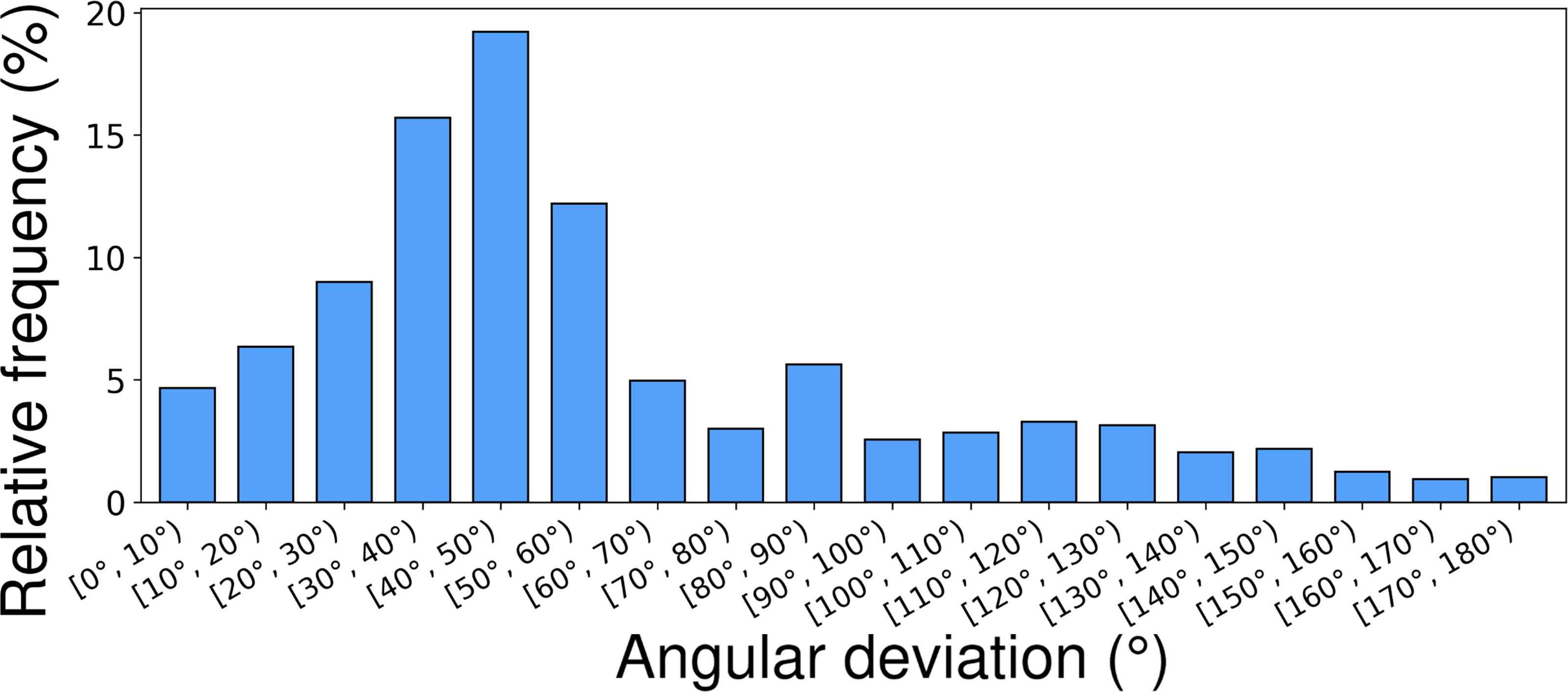}
\vspace{-1mm}
\caption{
Histogram of angular deviations (degrees) at the first frame after patch insertion ($K{+}1$) on LIBERO–Long,
showing the difference between predicted action directions with and without our adversarial patch.
Results are aggregated over all tasks in the Long suite.
}
    \label{fig:trajectory}
\end{figure}

\paragraph{\revised{Analysis of trajectory smoothness disruption}}
\revised{We evaluate the trajectory term by measuring the angular deviation between the action directions predicted under attack and in the benign rollout at the first frame after patch application ($K{+}1$), as shown in Fig.~\ref{fig:trajectory}.
Most deviations concentrate in the $30^\circ$--$60^\circ$ range, indicating a substantial direction change as soon as the patch takes effect.
Such deviations steer the end-effector off its nominal path into high-curvature motions, supporting $\mathcal{L}_{traj}$ as a curvature-oriented surrogate for disruptive motion.
}

\paragraph{\revised{Effect of trajectory sampling and curvature estimation}}
\revised{
The trajectory term samples $U$ candidate future directions to find motions that increase curvature relative to the prefix.
In Tab.~\ref{tab:u_sensitivity}, performance improves substantially from $U{=}50$ to $U{=}100$, while larger values bring only marginal gains at higher cost; we thus use $U{=}100$.
Spline-based curvature estimation also stays reliable under short prefixes (Tab.~\ref{tab:spline_stability}): at $K{=}10$, fitting remains valid in all episodes, with high Top-5 agreement and a positive $\Delta\kappa$ rate above 70\%.
}

\begin{table}[!t]
\setlength\tabcolsep{4pt}
\centering
\caption{\revised{
Effect of candidate direction number $U$ at $K{=}30$, averaged over four LIBERO suites.
}}
\label{tab:u_sensitivity}
\vspace{-1mm}
\begin{tabular}{c|ccccc}
\hline
\revised{$U$} & \revised{50} & \revised{75} & \revised{100} & \revised{150} & \revised{200} \\ \hline
\revised{ASR} & \revised{44.6} & \revised{64.8} & \revised{80.1} & \revised{79.8} & \revised{\textbf{80.7}} \\
\revised{nASR} & \revised{72.1} & \revised{83.1} & \revised{\textbf{89.3}} & \revised{87.7} & \revised{88.7} \\ \hline
\end{tabular}
\vspace{-1mm}
\end{table}

\begin{table}[!t]
\setlength\tabcolsep{2pt}
\centering
\caption{\revised{
Spline-curvature stability under short prefix lengths with $U{=}100$, averaged over four LIBERO suites.
}}
\label{tab:spline_stability}
\vspace{-1mm}
\scalebox{1}{%
\begin{tabular}{c|ccc}
\hline
\revised{$K$} & \revised{10} & \revised{20} & \revised{30} \\ \hline
\revised{Top-5 agreement (\%) $\uparrow$} & \revised{$90.3{\pm}1.9$} & \revised{$95.0{\pm}1.9$} & \revised{$92.3{\pm}2.0$} \\
\revised{Positive $\Delta\kappa$ rate (\%) $\uparrow$} & \revised{$70.8{\pm}4.7$} & \revised{$90.7{\pm}2.8$} & \revised{$81.7{\pm}3.7$} \\ \hline
\end{tabular}
}
\vspace{-1mm}
\end{table}

\paragraph{\revised{Effectiveness on the Hume VLA}}
\revised{To verify that the attack is not restricted to OpenVLA, we evaluate it on Hume~\cite{song2025hume}.
Using the official models, Tab.~\ref{tab:hume} reports ASR at $K{=}10$ and $K{=}30$: the full method performs best across most settings, and removing either loss term reduces performance, confirming effectiveness on another VLA architecture.}


\begin{table}[!t]
\setlength\tabcolsep{2.4pt}
\setlength{\tabcolsep}{2.4pt}

\caption{
Attack success rate (ASR, \%) on official Hume models~\cite{song2025hume} trained on the four LIBERO suites, with prefix lengths $K\in\{10,30\}$.
\revised{Results are reported for the full method and variants without $\mathcal{L}_{sem}$ or $\mathcal{L}_{traj}$.}
}

\vspace{-1mm}
\label{tab:hume}
\begin{tabular}{c|cc|cc|cc|cc}
\hline
             & \multicolumn{2}{c|}{Spatial} & \multicolumn{2}{c|}{Object} & \multicolumn{2}{c|}{Goal} & \multicolumn{2}{c}{Long} \\ \cline{2-9} 
             & K=10          & K=30         & K=10         & K=30         & K=10        & K=30        & K=10        & K=30       \\ \hline
Ours w/o $\mathcal{L}_{sem}$   & 46.1 & 67.3 & 45.4 & 77.3 & 54.0 & \textbf{84.2} & 57.9 & 68.7 \\
Ours w/o $\mathcal{L}_{traj}$  & 38.7 & 74.2 & 48.0 & 69.5 & 52.1 & 78.2 & 56.6 & 73.0 \\
Ours                           & \textbf{52.2} & \textbf{86.3} & \textbf{56.4} & \textbf{82.2} & \textbf{62.6} & 83.8 & \textbf{69.0} & \textbf{84.5} \\ 
\hline
\end{tabular}
\vspace{-1mm}

\end{table}

\paragraph{\revised{Robustness against simple input-level defenses}}
\revised{We evaluate two input-level defenses at $K{=}30$, JPEG compression ($q{=}60$) and Gaussian blur ($r{=}2$). As shown in Tab.~\ref{tab:defense}, the average ASR stays close to the undefended case (79.2/80.2 vs. 81.0), so these simple defenses alone cannot suppress the attack.}

\begin{table}[!t]
\setlength\tabcolsep{3.4pt}
\centering
\caption{\revised{
ASR (\%) under input preprocessing defenses at $K{=}30$.
JPEG compression and Gaussian blur are applied to each patched observation before VLA inference.
}}
\label{tab:defense}
\vspace{-1mm}
\begin{tabular}{l|cccc|c}
\hline
\revised{Defense} & \revised{Spatial} & \revised{Object} & \revised{Goal} & \revised{Long} & \revised{AVG} \\ \hline
\revised{JPEG ($q{=}60$)} & \revised{70.6} & \revised{89.1} & \revised{69.4} & \revised{\textbf{87.5}} & \revised{79.2} \\
\revised{Gaussian blur ($r{=}2$)} & \revised{\textbf{74.1}} & \revised{86.2} & \revised{\textbf{75.3}} & \revised{85.2} & \revised{80.2} \\
\revised{No Defense} & \revised{73.8} & \revised{\textbf{90.7}} & \revised{72.8} & \revised{86.6} & \revised{\textbf{81.0}} \\ \hline
\end{tabular}
\vspace{-1mm}
\end{table}



\paragraph{\revised{Failure-case analysis}}
\revised{
Several conditions reduce the attack's impact.
Very short prefixes weaken it: at $K{=}10$ the prefix gives limited context for localization and curvature estimation, and ASR drops markedly (e.g., 33.4\% on Spatial and 47.0\% on Goal in Tab.~\ref{tab:asr}).
The Spatial suite is also harder to disrupt, as its short, tightly constrained motions let small induced deviations be corrected before completion.
On shorter-horizon suites a large prefix leaves only a brief suffix to influence, so larger $K$ does not always help (e.g., the slight ASR drop from $K{=}30$ to $K{=}40$).
The attack is thus most effective with a moderate prefix on longer-horizon tasks.}

\begin{figure}[!t]
    \centering
    \includegraphics[width=1\linewidth]{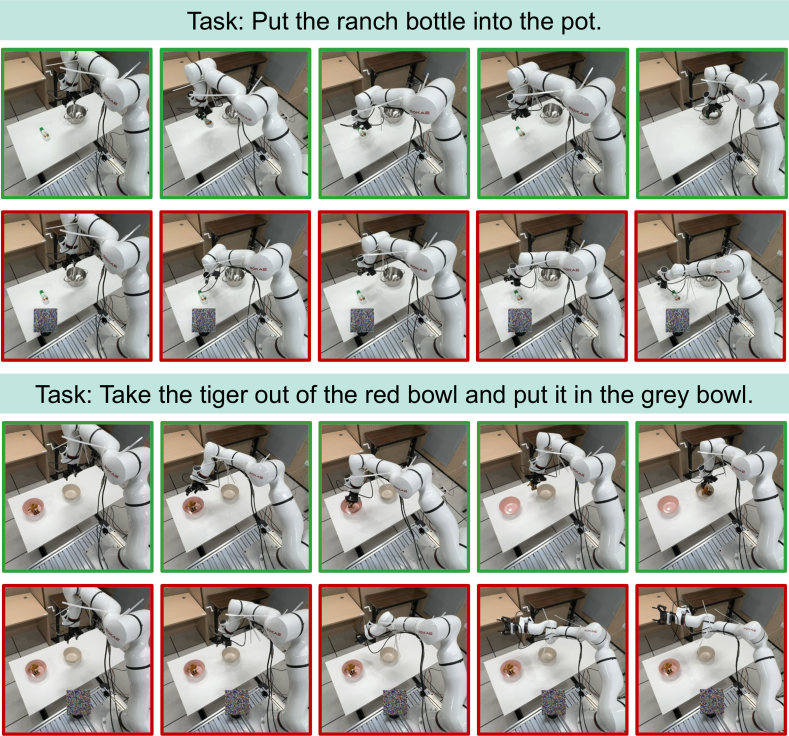}

    \vspace{-1mm}
\caption{
Snapshots of real-world VLA task executions on a ROKAE xMate ER7 Pro robot.
For each task, the top row shows benign execution and the bottom row shows execution with a physically printed adversarial patch mounted front-facing in the workspace.
}
    \label{fig:physics}
\end{figure}



\paragraph{Physical attack evaluation}
We evaluate the attack on a ROKAE xMate ER7 Pro arm with a Robotiq 2F-85 gripper, using an overhead Intel RealSense D455 RGB camera and OpenVLA-7B for policy execution. 
The adversarial patch is physically printed at 18\,cm $\times$ 18\,cm and mounted on a rigid support in the workspace, keeping it approximately front-facing to the camera. 
The patch is captured as part of the physical scene and affects the VLA through real visual observations. 
Fig.~\ref{fig:physics} shows representative attacked executions, where the robot exhibits incorrect object interactions and fails to complete the tasks.

We further report quantitative results on two real-world manipulation tasks, each repeated for 50 trials under benign and attacked conditions. 
As shown in Tab.~\ref{tab:physical_quant}, the printed patch reduces success rates from 72\% to 12\% on Task A and from 58\% to 8\% on Task B, confirming substantial degradation in physical execution.

\begin{table}[!t]
\setlength\tabcolsep{6pt}
\centering
\caption{\revised{
Quantitative physical evaluation. Each task is repeated for 50 trials under benign and attacked conditions.
Values report task success rate (\%).
Task A: put the ranch bottle into the pot.
Task B: move the tiger from the red bowl to the gray bowl.
}}
\label{tab:physical_quant}
\vspace{-1mm}
\begin{tabular}{c|cc}
\hline
\revised{Condition} & \revised{Task A} & \revised{Task B} \\ \hline
\revised{Benign} & \revised{72} & \revised{58} \\
\revised{Attacked} & \revised{12} & \revised{8} \\ \hline
\end{tabular}
\vspace{-2mm}
\end{table}

%% file: 7-conclusion.tex
\section{Conclusion}

In this paper, we have presented a framework for adversarial patch attacks on VLA-based robots under partial observability.  
Even when only a limited prefix of the trajectory is available, exploiting vulnerabilities in perception and control can induce persistent disruptions.  
Extensive experiments in simulation and the real world show that our method causes long-horizon failures and significantly reduces task success rates, underscoring the risks of adversarial threats to VLA models and motivating systematic evaluation and robustness enhancement for VLA-based robotic systems.


\noindent\revised{\textbf{Limitations and Future Work.}
Our partially observable setting is weaker than full-rollout optimization, but still assumes partial-access gray-box information and a short execution prefix.
The method relies on cross-modal attention or similar grounding signals, as well as prefix trajectory information, and therefore does not cover fully black-box policies.
Extending the attack to weaker-access settings and broader VLA architectures remains an important direction for future work.}